# AN ONTOLOGY-BASED APPROACH TO DATA EXCHANGES FOR ROBOT NAVIGATION ON CONSTRUCTION SITES

***Sina Karimi;***
***Construction Engineering Department, École de Technologie Supérieure (ÉTS), Montréal, Québec, Canada;***
***sina.karimi.1@ens.etsmtl.ca***

***Ivanka Iordanova;***
***Construction Engineering Department, École de Technologie Supérieure (ÉTS), Montréal, Québec, Canada;***
***ivanka.iordanova@etsmtl.ca***

***David St-Onge;***
***Mechanical Engineering Department, École de Technologie Supérieure (ÉTS), Montréal, Québec, Canada;***
***david.st-onge@etsmtl.ca***

***SUMMARY:*** *With growth in the use of autonomous Unmanned Ground Vehicle (UGV) for automated data collection from construction projects, the problem of inter-disciplinary semantic data sharing and exchanges between construction and robotic domains has attracted construction stakeholders' attention. Cross-domain data translation requires detailed specifications especially when it comes to semantic data translation. Building Information Modeling (BIM) and Geographic Information System (GIS) are the two technologies to capture and store construction data for indoor structure and outdoor environment respectively. In the absence of a standard format for data exchanges between the construction and robotic domains, the tools of both industries are yet to be integrated in a coherent deployment infrastructure. Hence, the semantics of BIM-GIS cannot be automatically integrated by any robotic platform. To enable semantic data transfer across domains, semantic web technology has been widely used in multidisciplinary areas for interoperability. We exploit it to pave the way to a smarter, quicker and more precise robot navigation on construction sites. This paper develops a semantic web ontology integrating robot navigation and data collection to convey the meanings from BIM-GIS to the robot. The proposed Building Information Robotic System (BIRS) provides construction data that are semantically transferred to the robotic platform and can be used by the robot navigation software stack on construction sites. To reach this objective, we first need to bridge the knowledge representation between construction and robotic domains. Then, we develop a semantic database to integrate with Robot Operating System (ROS) which can communicate with the robot and the navigation system in order to provide the robot with semantic building data at each step of data collection. Finally, the proposed system is validated through a case study.*

## 1. INTRODUCTION

Progress monitoring of construction projects needs accurate and comprehensive data collection to increase productivity and support risk management. Manual data collection is inefficient due to its high cost, inaccuracy, and error-prone nature (Rebolj et al., 2017). Among different propositions such as implementation of new technologies by leveraging cross-functional teams (Barbosa et al., 2017) and Scrum strategy deployment in various phases of construction (Streule et al., 2016), the use of autonomous robots has shown great potential to achieve efficiency and high data collection precision (Ardiny et al., 2015). Studies have shown that the application of mobile robots on construction sites for progress monitoring is one of the most important ones (Albeaino and Gheisari, 2021). As Albeaino et al. (2019) argue, Unmanned Aerial Vehicles (UAV) acquire construction data using commercial sensors such as laser scanners, cameras, radio frequency identification, and ultrasonic beacon systems. While UAV are mainly use for outdoor data acquisition, UGV complement the application of UAV for indoor environments where Global Positioning System (GPS) is limited (Lin and Golparvar-Fard, 2020). Another approach adopted by Asadi et al. (2020) use an integrated UAV-UGV system for construction data acquisition where UGV capabilities are limited due to cluttered nature of construction sites with many obstacles. In this direction, the authors leveraged UAV to address UGV limitations to provide comprehensive and accurate data (Asadi et al., 2020). The fundamental requirement of automated data collection is for the robot to be able to (1) navigate autonomously in a dynamic environment and (2) acquire relevant and accurate data. With the growth of mobile robots' capabilities in recent years, the interest in having high-level semantic information integrated within the robot is increasing: it is expected to ultimately make the robots easier to deploy (Crespo et al., 2020). Robots with semantic representation and recognition (association of location-semantic) of their environment are more intuitive to operate for non-experts, because they share conceptual understanding (Kostavelis and Gasteratos, 2017). Among the various challenges to implement semantic navigation, one is the association of high-level information with the geometry of the environment, such as discrete maps often referred to as occupancy grids. To achieve this, one needs to develop a knowledge representation (Crespo et al., 2020), for which domain-specific ontologies are well suited (Gruber, 1995). Therefore, this study intends to facilitate data exchange between construction and robotic domains by developing an ontology which supports formal representation of the domains' knowledge. The resulting ontology then becomes a base structure for data extraction of BIM-GIS; relevant to robot navigation. The current study is part of a larger research project aiming at automated data collection and progress monitoring of construction projects.

BIM as a product is used throughout the entire life cycle of a facility in many aspects including design to maintenance state tracking (Doumbouya et al., 2017), smart city integration (Afsari et al., 2019), 4D simulation (Hatori et al., 2020), and clash detection (Savitri and Pramudya, 2020). In a similar way, GIS has enabled Architecture, Engineering, Construction (AEC) industry to acquire geo-spatial data with regards to surrounding site conditions and topographical information of a construction site (Karan and Irizarry, 2015). The integration of BIM and GIS grants new opportunities in construction projects such as finding the optimal location for tower cranes (Irizary and Karan, 2012), assessing the occupants' behavioral impact on energy consumption (Afkhamiaghda et al., 2019), assessing urban energy performance (Yamamura et al., 2017), helping with emergency response route (Tashakkori et al., 2015), merging indoor-outdoor combined route planning (Teo and Cho, 2016) and pre-construction planning (Karan and Irizarry, 2015). Integration of BIM-GIS data provides a holistic overview of digital built environment including the facility and the environment around it at urban scale that can be used for robot navigation on construction sites for data acquisition of the existing condition. However, the construction and robotic domains need to share a common semantic interoperability.

Semantic web technologies provide high-level data exchange among various domain knowledge representations, enabling interoperability through attachment of decentralized data with semantics to different concepts (Kalfoglou, 2009). In an ontology, concepts, semantics and their relations with one another are defined through taxonomies (hierarchical structure of data) and relationships (Van Rees, 2003), thereby providing a machine-understandable structure in which concepts and knowledge are represented (El-Diraby et al., 2005). Ontologies can be categorized into three levels based on their components and level of detail, namely: top-level (or upper), domain and application ontologies. Top-level ontologies formalize a generic ontology across all domains whereas domain ontologies provide knowledge representation that is formal, reusable, and shareable across a specific domain. Application ontologies aim to create focused ontology for a given application (Karan Ebrahim P. et al., 2016).

The Institute of Electrical and Electronics Engineers published a standard (IEEE 1872-2015) ("IEEE Standard Ontologies for Robotics and Automation," 2015) for Robotics and Automation ontologies based on the Core Ontology for Robotics and Automation (CORA). In this paper, we leverage IEEE 1872-2015 in order to bridge it with construction-based knowledge. This standard provides formal and shareable representation of the robotics and automation domain along with the definition of concepts and the relationships between concepts, attributions and constraints. The proposed ontology is a domain ontology that needs to be derived into specific implementations. Since some of the concepts discussed in the CORA are not thoroughly defined, CORAX (CORA's extension) was proposed to cover these gaps ("IEEE Standard Ontologies for Robotics and Automation," 2015). The other standard we leveraged is the IEEE 1873-2015 ("IEEE Standard for Robot Map Data Representation for Navigation," 2015). Robot Map Data Representation for Navigation (MDR) standard focuses on the interoperability between robots, humans, and machines in terms of 2D metric and topological maps rather than 3D or semantic maps. Hence, the current study intends to bridge the semantic information of BIM and GIS using some of the concepts in MDR in order to enrich map representation for human-robot data exchange.

Despite the number of studies using mobile robot platform to acquire data of construction projects, semantic data translation between construction and robotic domains is still a problem requiring greater attention for research. In this direction, since there is no standard format to translate data between BIM-GIS and Robot Operating System (ROS), the semantics of the IFC and CityGML is not understood by the robotic platform, thereby making it essential for the robot to be able to “call” for the semantic information (Karan and Irizarry, 2015; Meschini et al., 2016). The current study’s objective is to enable semantic interoperability between BIM, GIS, and robot navigation system to translate semantic data from BIM and GIS to the ROS. Building elements' semantics can be provided to the ROS so that different information would be accessible to the robot when and where it is needed. The proposed bridging ontology is based on the extension of IEEE 1872-2015 and IEEE 1873-2015 standards. To reach this objective, we develop a Building Information Robotic System (BIRS) which conveys semantic data to ROS. The current study’s contributions are:

- An ontology bridging BIM-GIS data to IEEE 1872-2015 and IEEE 1873-2015 standards.
- Cross-domain data structure enabling exchanges between construction and robotic domains.
- A practical implementation of the proposed ontology and data structure to deploy it on an autonomous mobile robot navigating a construction site.

The first contribution of this paper is a theoretical contribution providing a conceptual solution to bridge the knowledge representation of construction and robotic domains while the second and the third are the practical contributions providing the means for deploying the first contribution.

The existing robot navigation solutions are categorized as “heuristic-based” and “classical” methods (Wahab et al., 2020). Global path planners such as Potential Field, Cell Decomposition, and Dijkstra require the robot to have a priori understanding of the environment with the obstacles’ locations (Hassan et al., 2017). Potential Field constantly searches locations of the given graph to find the target (Liu et al., 2006). Cell Decomposition algorithm such as A* divides the mobile robot environment into cells which finds the shortest distance between the start and the target connected through the cell graphs (Esan et al., 2020). Dijkstra algorithm is also a graph search-based method from the starting point to the target connected by line (Gul et al., 2019). Local path planners, on the other hand, such as Vector Field Histogram and Genetic Algorithm create a new path in response to changes in the robot environment (Esan et al., 2020). Vector Field Histogram detects and avoids obstacles on the way to the target and is not considered appropriate for dynamic environments (Almasri et al., 2016). Genetic Algorithm adopts problem-solving approach by addressing problems when they occur through several iterations (Wang et al., 2016). Aligning the state-of-the-art navigation solutions, this paper provides waypoints for transiting from one space to another optimizing the existing navigation solutions.

The remainder of this paper is structured as follows: Section 2 describes other scientific works related to the studied domain to provide a background of what has been achieved so far as well as their contribution. Section 3 provides a detailed explanation of the methodology and how our ontology is developed to facilitate data exchange. Section 4 presents how the ontology can be used as well as the results from a case study. Finally, section 5 concludes the research carried out with the limitations of the proposed system as well as some future research directions.

## 2. RELATED WORK

Even though the growth of World Wide Web has resulted in incredible increase of the information coming from heterogeneous data sources, users are able to navigate through the information easily. This implies the power of semantic web in a way that is understandable by humans and machines (Berners-Lee and Hendler, 2001) for a domain-specific shared knowledge (Wang and Issa, 2020). The implementation of semantic web requires formal knowledge representation of domains (ontologies) (Wang et al., 2019) in which the concepts and the relationships between them are explicitly described (Usmani et al., 2020). Ontology is defined in different ways in various domains. Guarino (1995) defined ontology as *"an artefact constituted by a specific vocabulary used to describe a certain reality, along with a set of explicit assumptions related to the desired meaning of the vocabulary”* in the context of Artificial Intelligence. Venugopal et al. (2015) defined ontology as a *“knowledge representation* *mechanism"* which provides an overview of a specific domain at abstract level. Studer et al. (1998) defined ontology at a higher level: *''a formal, explicit specification of a shared conceptualization"*. Pauwels et al. (2017), on the contrary, defined ontology and its applications at a lower level by *"enabling data integration and complex search queries across several data sources."* When it comes to developing an ontology, vocabulary does not suffice to convey the intended meaning. In order to make an ontology efficient and functional, other parameters are taken into account, namely the concepts (terms which are mainly abstract and are aligned to the taxonomies), relationships (the semantic connection between concepts), instances (an existing entity representing features of concepts) and axioms (defined rules across the domain which is valid) (González et al., 2020). The potentials of ontologies have resulted in the increase of interest in using them and therefore, novel forms of knowledge representation are created. In this direction, the World Wide Web Consortium (W3C) developed the Resource Description Framework (RDF) to function as a *''standard model for data interchange on the Web"* (World Wide Web Consortium, 2014).

In the construction industry, the earliest studies on semantic web technologies aimed at achieving a higher level of efficiency for the federated information through adopting ontological approach to retrieve the *"related concepts"* to manage on-site problems (Elghamrawy and Boukamp, 2008). Another study used ontology to search and extract the construction data because the AEC data are numerous and somewhat hard to retrieve (Staub-French and Nepal, 2007). Sensing data integration for construction management through semantic web was also subject to several earlier contributions. Elghamrawy and Boukamp (2010) developed an ontology in which Radio-frequency identification (RFID) technology is used to archive and retrieve construction document information. Liu et al. (2016) used ontology for cost estimation of construction projects in China. Barbau et al. (2012) developed a plugin for a software tool supporting ontology development (i.e. Protégé) to translate the STEP's EXPRESS schema into Web Ontology Language (OWL). Farias et al. (2015) proposed a framework to create an OWL ontology for COBie, a standard in BIM for Facility Management. González et al. (2020) described the development of an ontology to support indoor navigation which is based on IfcOWL ontology translating IFC schema into OWL. Venugopal et al. (2015) developed a model for data exchanges using IFC schema based on ontologies. Ontological approaches were also suggested for information exchanges in precast concrete (Venugopal et al., 2012). The literature shows that ontologies are nowadays common for information exchanges across the AEC industry but they do not fully take advantage of established ontologies in other domains such as robotics. Therefore, we propose a bridging ontology between construction and robotic domains to cover this gap.

Many other fields have leveraged ontologies to extend interoperability across other domains. In the AEC industry, the nature of the information is decentralized and fragmented (Atazadeh et al., 2017) which results in obstruction for interoperability purpose. The data from different sources come with different formats and they follow different ontologies. Each of those ontologies are designed for different needs, therefore, it may obstruct interoperability in the AEC industry (Aziz et al., 2006). Hence, there has been a considerable number of attempts to provide open source data schema for BIM interoperability such as BIMXML and COINS (Zhu et al., 2018). However, Industry Foundation Classes (IFC) is the primary open-source, EXPRESS-based data schema being used across the AEC domain (Mignard and Nicolle, 2014). IFC schema is comprised of four layers, namely: resource layer, core layer, interoperability layer and domain/application layer (Terkaj and Viganò, 2017). IFC is a hierarchical data schema in which classes inherit the properties of upper layers (González et al., 2020). Furthermore, many papers address the challenges of BIM and GIS integration in which some studies adopt ontological approach. More information in this regard and the requirements for utilizing BIM-GIS integration in robot navigation can be found in the review by Karimi and Iordanova (2021).

There is also a number of ontology-based deployments in robotics outside of the construction domain. One of the first studies which used ontology is KnowRob (Knowledge processing for Robots) (Tenorth and Beetz, 2009). KnowRob ontology provides an open-source knowledge system for service robots managing uncertainties such as sensors' noise. In KnowRob ontology, the authors argue that controlling an autonomous robot requires several factors such as representing more fine-grained action. In another study (Beetz et al., 2018), the authors extended the KnowRob ontology to retrieve experimental knowledge (*"narrative enabled episodic memory"*). OpenEASE is a service, based on the web knowledge using KnowRob to retrieve, store, supervise, and visualize the robot knowledge (Beetz et al., 2015). KnowRob also uses RDF to represent the knowledge. RObot control for Skilled ExecuTion of Tasks in natural interaction with humans; based on Autonomy, cumulative knowledge and learning (ROSETTA) used a set of ontologies to constitute a model in order for the manufacturing robots to be adopted (Olivares-Alarcos et al., 2019). ROSETTA ontology is based on CORA ontology, which is in turn based on SUMO ontology. Although CORA is using SUO-KIF language, ROSETTA is written in OWL. Persson et al. (2010) developed a *"knowledge integration Framework"* which establishes relationships between different segments of ROSETTA. OROSU (Ontology for Robotic Orthopedic Surgery) is yet another ontology developed based on IEEE 1872-2015 Standard Ontology for Robotics and Automation, uses OWL to retrieve information of the knowledge-based framework for surgical robotics (González et al., 2020). Diab et al. (2019) proposed Perception and Manipulation Knowledge (PMK) ontology for representing and reasoning knowledge in task and motion planning. In PMK ontology, the authors implemented from OWL the ontology using again CORA and SUMO as the base ontologies (Olivares-Alarcos et al., 2019).

In the sub-domain of semantic knowledge for robot navigation, Kollar and Roy (2009) adopted an ontological approach to enable human-robot interaction, specific to the task of search and find. Galindo et al. (2005) proposed a multi-hierarchical approach for semantic navigation comprising of spatial and conceptual hierarchies. The former describes conventional metric approach of the building spaces and the latter incorporates semantic information of the environment. However, the BIM semantics still remains neglected even if it can add more semantic information to robot navigation. With great contribution to semantic navigation, none of them studied the necessity of incorporating BIM semantics to robot navigation during and/or after construction phase. There are also already several works addressing BIM usage for robot navigation. Delbrügger et al. (2017) developed a framework for digital twin factories supporting human and robot indoor navigation. Ibrahim et al. (2017) studied the use of aerial robots to capture data from construction sites in an interactive way. Another study examined the use of BIM for localization of Unmanned Aerial Vehicle (UAV) in indoor environment taking advantage of AprilTags (Nahangi et al., 2018). Siemiątkowska et al. (2013) studied the use of BIM-based map representation in which the robot could localize semantically using hierarchical path planning. HAMIEH et al. (2017) developed a four step BIM-based path planning strategy which uses hierarchical refinement of the number of paths. Palacz et al. (2019) proposed graph-based approach for indoor navigation using BIM/IFC. Despite their great contributions, none of the aforementioned works studied the automatic translation of semantic data from BIM to the ROS. There also have been several research attempts in which the authors used GIS for robot navigation (Mirats-Tur et al., 2009; Qiangrong Yang et al., 2015; Yan et al., 2013). However, the integration with BIM and the additional information that can be used for robot navigation is yet to be thoroughly studied.

## 3. RESEARCH METHODOLOGY

The current study proposes a novel approach which BIM and GIS data are used in robot navigation during/after construction phase. The research methodology is comprised of two steps: (1) developing Building Information Robotic System (BIRS) Ontology in order to bridge IFC and CityGML schemas to IEEE 1872-2015 and IEEE 1873-2015, and (2) enabling cross-domain interoperability through BIRS Data Exchange. Figure 1 illustrates the pipeline establishing the practical implementation of the robotic system using BIM and GIS for navigation.

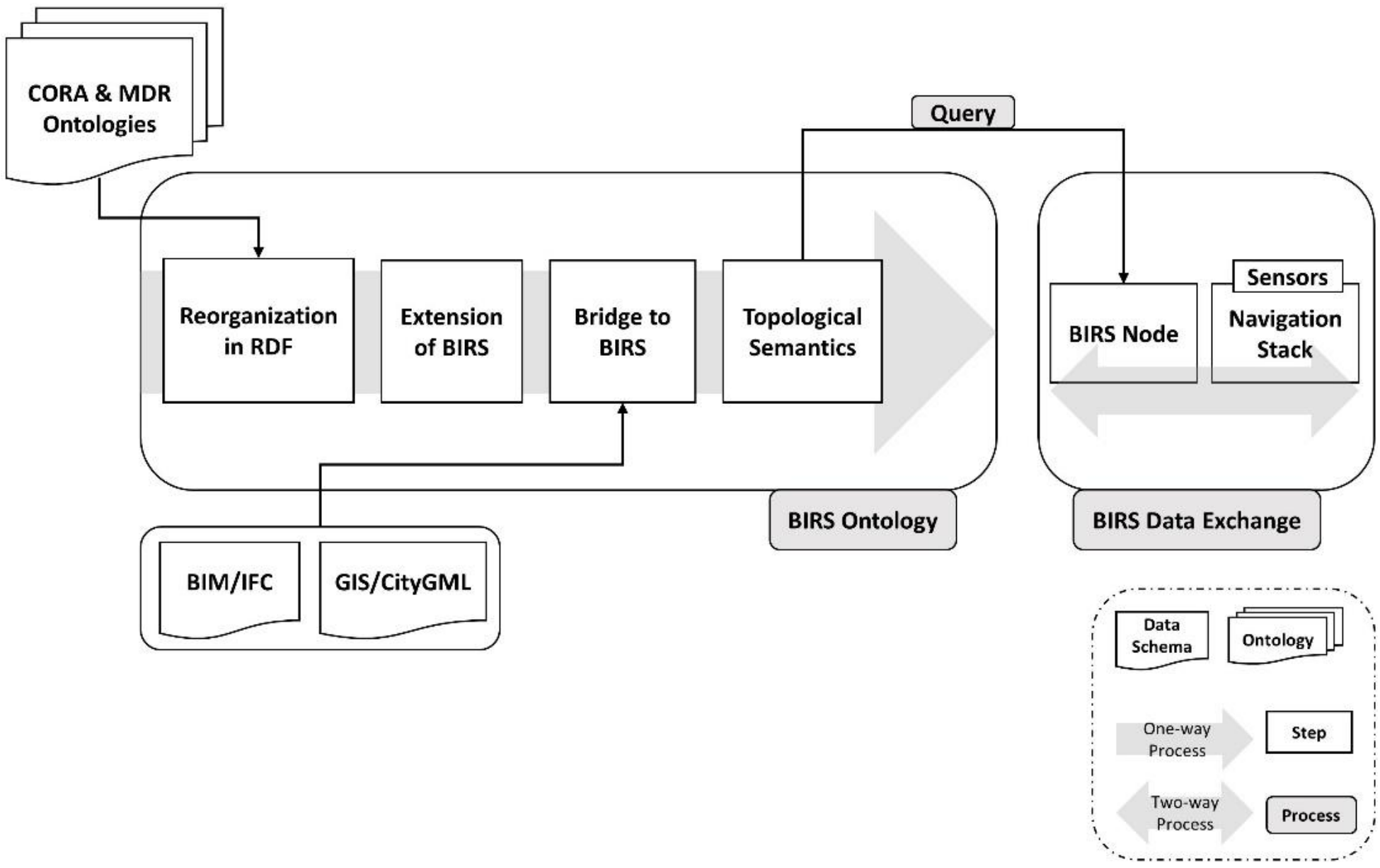

*FIG. 1. Proposed pipeline for implementing Building Information Robotic System (BIRS) supporting development of BIRS Ontology and BIRS Data Exchange.*

As illustrated in FIG. 1, the first step (BIRS Ontology) creates an ontology bridging IFC and CityGML data to CORA and MDR. The second step (BIRS Data Exchange) reasons the BIRS ontology to create a BIRS node in ROS in order to translate the semantic data from the ontology to the robotic system. Both steps in the pipeline are explained in detail in the following sections.

## 3.1.Building Information Robotic System (BIRS) Ontology

The purpose of this section is to develop an ontology supporting BIM and GIS bridging to CORA and MDR. To achieve this, the existing IEEE standards need to be reorganized (the ontology classes and axioms remain intact) to facilitate the knowledge translation from engineering to application ontologies. Then, we express CORA and MDR in the RDF to be compatible with the development of median level classes. The combination of RDF and XML would enable users to interchange data among various applications (Karan and Irizarry, 2015). Since the resulting entities should comply with IEEE 1872-2015 and IEEE 1873-2015 standards, the intermediate level classes are developed to be integrated with the correspondent concepts from IFC and CityGML.

In order to extend CORA and MDR in our work, a short recall of the underlying concepts is required. CORA extends the Suggested Upper Merged Ontology (SUMO) which is a top-level ontology (Prestes et al., 2013). SUMO supports definition of high-level knowledge concepts in the world. The highest SUMO class is *Entity* which has two further sub-classes; *Physical* and *Abstract*. The former describes the entities in 3D space and the latter represents the concepts which do not have spatial shape. In other words, *Abstract* contains anything which does not fall into *Physical* (Niles and Pease, 2001). *Physical* is a disjoint partition of *Object* and *Process*. *Object* describes existent objects in 3D space with no temporal effect on the space while *Process* follows a *"perdurantist"* approach that adds temporal effects and considers 4D orientation (Fiorini et al., 2015). *Abstract*, on the other hand, is defined not to be a subclass of *Object* and is categorized into the following sub-classes: *Quantity*, *Attribute*, *SetOrClass*, *Relation*, and *Proposition*. Figure 2 illustrates the SUMO taxonomies on which the CORA is developed.

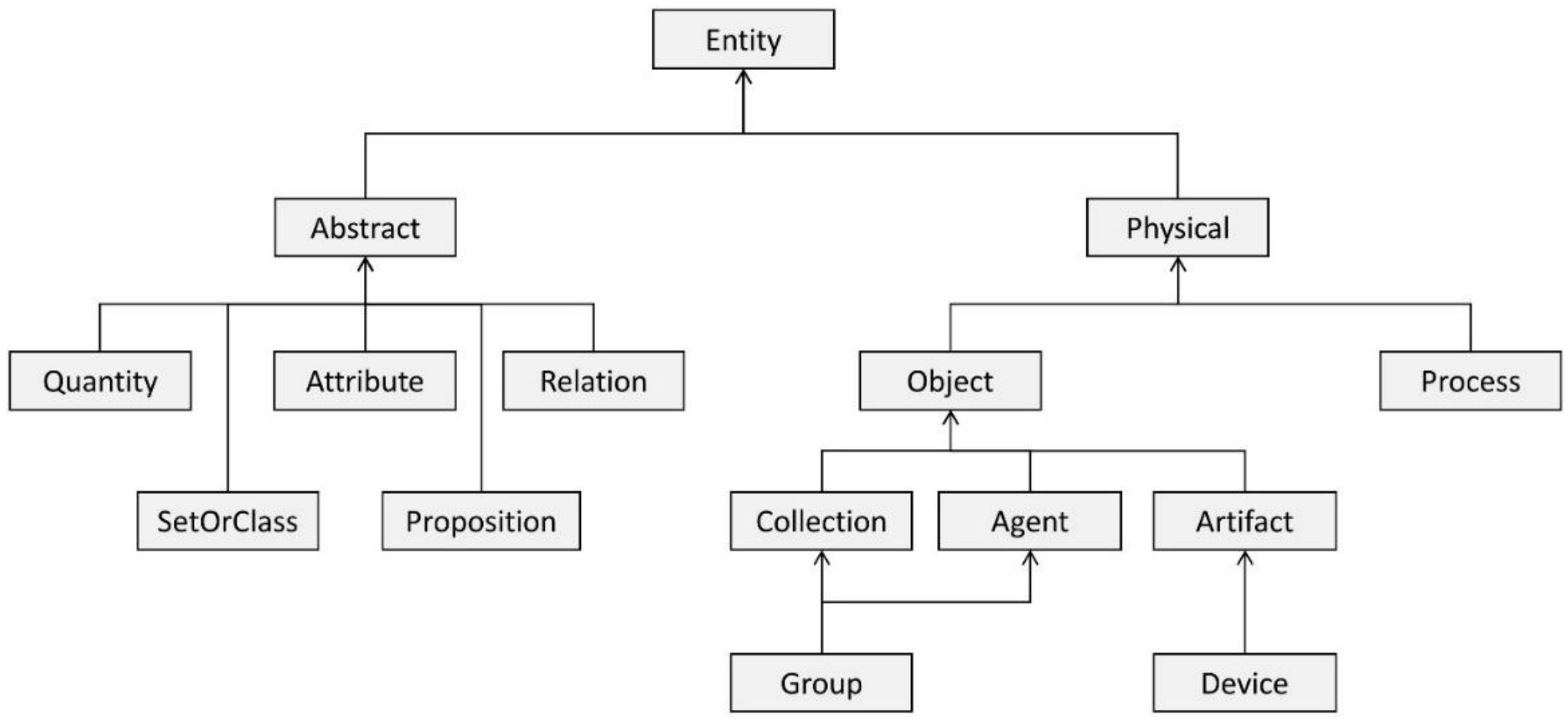

*FIG. 2. Basic SUMO taxonomies [Source: (“IEEE Standard for Robot Map Data Representation for Navigation,” 2015)]*

IEEE 1872-2015 argues that *"Concepts and relations associated with design, interaction, and environment"* are too general to be incorporated in CORA and are not effectively covered by SUMO (“IEEE Standard Ontologies for Robotics and Automation,” 2015). To cover this gap, Fiorini et al. (2015) defined a new class *Physical Environment* as a three dimensional environment which contains region and other artifacts existing in 3-D space dependent on the landmark. CORAX incorporates *PhysicalEnvironment* as the subclass of *Object* under *Physical* which is located in 3 dimensions and consist of one *Region* in minimum (“IEEE Standard Ontologies for Robotics and Automation,” 2015). IEEE 1872-2015 also defines *Design* as a subclass of *proposition* which has idealization relationship between design and artifact. In SUMO ontology, *content bearing physicals* represent propositions such as a descriptive sentence, a graph, etc. No restriction is applied for *ContentBearingPhysicals* to represent an idea. Therefore, we integrate the MDR as the subclass of *CORA:ContentBearingObject* since MDR aims to facilitate data interoperability between robots. IEEE 1873-2015 standard (“IEEE Standard for Robot Map Data Representation for Navigation,” 2015) subdivides maps into *Metric* and *Topological* in which *Metric* maps are the disjoint entity of the *Continuous Metric Maps* and *Discrete Metric Maps* representing physical layout of the environment and the objects within the robot’s physical environment. *Continuous* metric maps are comprised of the geometric elements while *Discrete* metric maps utilize bitmap illustration of the environment under which *OccupancyGridMaps* are categorized. *OccupancyGridMaps* are the most widely used maps in Simultaneous Localization and Mapping (SLAM) and generally in robot navigation; are considered as *Discrete Metric Maps*. *Topological* maps are generally represented by sets of nodes and edges (Choi et al., 2011) which facilitates path planning task with running an algorithm on the created graph (“IEEE Standard for Robot Map Data Representation for Navigation,” 2015). FIG. 3 illustrates the overall integral graph in which all the ontologies come together and shape an integral ontology taxonomy along with the proposed BIRS entities.

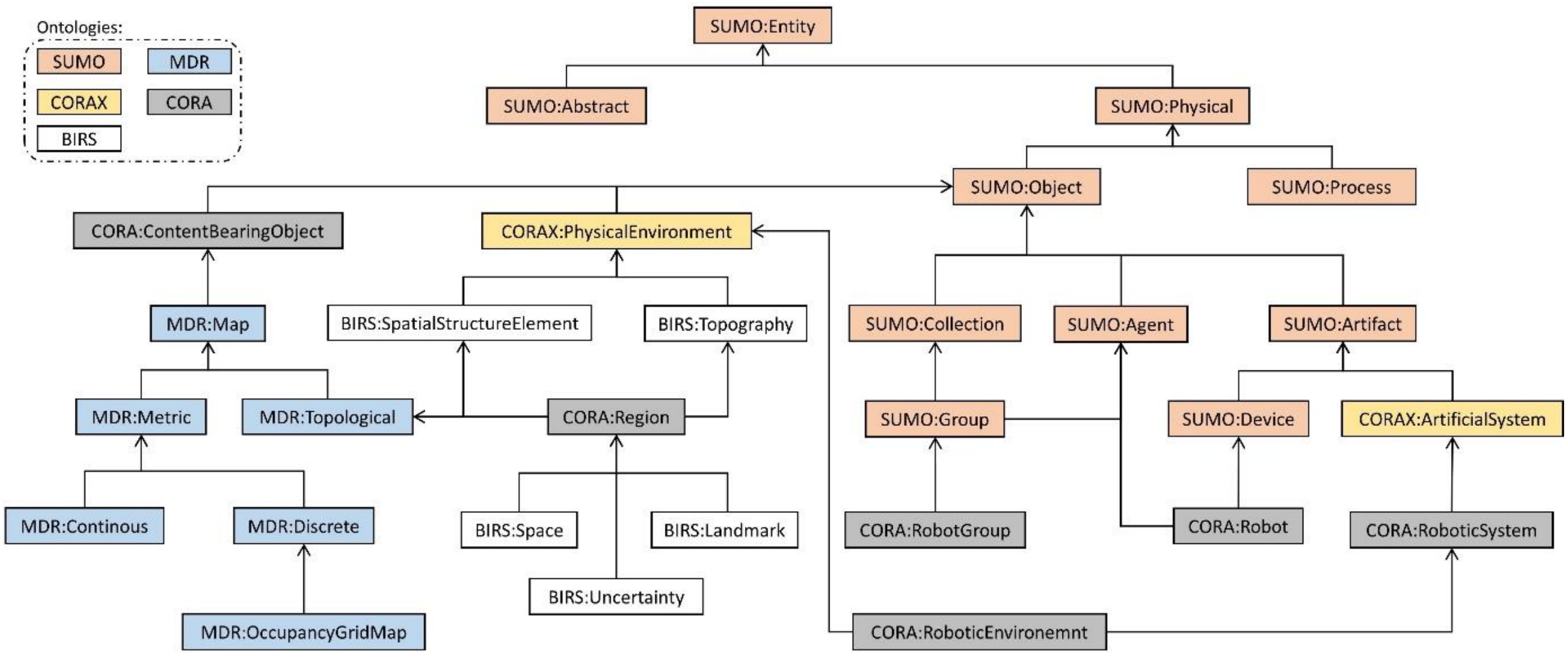

*FIG. 3. Integration of CORA and MDR ontologies with Building Information Robotic System (BIRS) Ontology indicating sub-class relationships (:SubClassOf) [Sources: ("IEEE Standard for Robot Map Data Representation for Navigation," 2015; "IEEE Standard Ontologies for Robotics and Automation," 2015)]*

We extend an intermediate level between *CORAX:PhysicalEnvironment* and *CORA:Region* to differentiate between indoor and outdoor environments. This is required by the digital representation of the built environment that correlates with building components and their semantic attributes with the topographic and the existing condition information of the construction site (Karan and Irizarry, 2015). Therefore, *SpatialStructureElement* and *Topography* can be the median level which define *Region* more effectively. We follow the extension of CORA and MDR by including *Landmark*, *Space* and *Uncertainty* as the sub-classes of *Region* bridging construction physical environment comprised of BIM and GIS. TABLE 1 provides the definition of the new concepts provided in the BIRS.

*TABLE 1. BIRS concepts definition.*

| *Concept* | *Definition* | *Reference* |
|---|---|---|
| *Landmark* | *"A feature that is used for localization of a mobile robot. It is a feature whose pose in a coordinate system can possibly be measured (localized) by the robot's sensors with respect to a given map."* | ("IEEE Standard for Robot Map Data Representation for Navigation," 2015) |
| *Space* | *" represents an area or volume bounded actually or theoretically. Spaces are areas or volumes that provide for certain functions within a building."* | *(buildingSMART, 2013)* |
| *Uncertainty* | Dynamics of the construction site which cannot be considered as the landmarks, and has temporal effects on the construction job-site. | - |

*Landmarks* are the physical building elements which the robot can use for localization. *IfcBuildingElement* is the entity which defines the properties of the building elements. Semantics of classes and properties are defined through axioms in the proposed BIRS ontology. The *SubClassOf* axiom is a demonstration of the defined relationship between the higher and the lower entity (Karan Ebrahim P. et al., 2016). As an instance, since the *IfcWall* is a subclass of *IfcBuildingElement*, *IfcWall* inherits all the properties of *IfcBuildingElement*. It is also applicable to all IFC classes due to the fact that IFC is a hierarchical data schema. In BIRS, the building elements extracted as the sub-class of *IfcBuildingElement* are *IfcWall*, *IfcCurtainWall*, *IfcColumn*, *IfcDoor*, *IfcRailing*, and *IfcStair*. FIG. 4 illustrates the relationships of BuildingElement with higher entity (landmark) and the properties associated with it in IFC schema. There are several quite essential properties in *IfcBuildingElement* for the translation of a building layout into a ROS compatible format. The shape and the location of the *landmark* can be derived through *ObjectPlacement* and *Material* properties. *IfcProduct* is the super-type of the *IfcBuildingElement* through which the local placement is defined. The coordinates of the geometric representation of *BuildingElement* is defined through *LocalPlacement*. Using the material properties of the *landmark* would be a contribution to robot navigation as well, but accessible through another pipeline than layout. Using the aforementioned IFC classes in the BIRS ontology, a BIM-based occupancy grid map is within reach. This map can be extracted in two ways. One is to use tools such as ifcConverter in order to export the map from IFC files to .svg, then to a ROS-compatible format such as .png, and the second method would use the BIM design authoring tools such as Autodesk® Revit and Graphisoft® ArchiCAD in order to export the map in .png format.

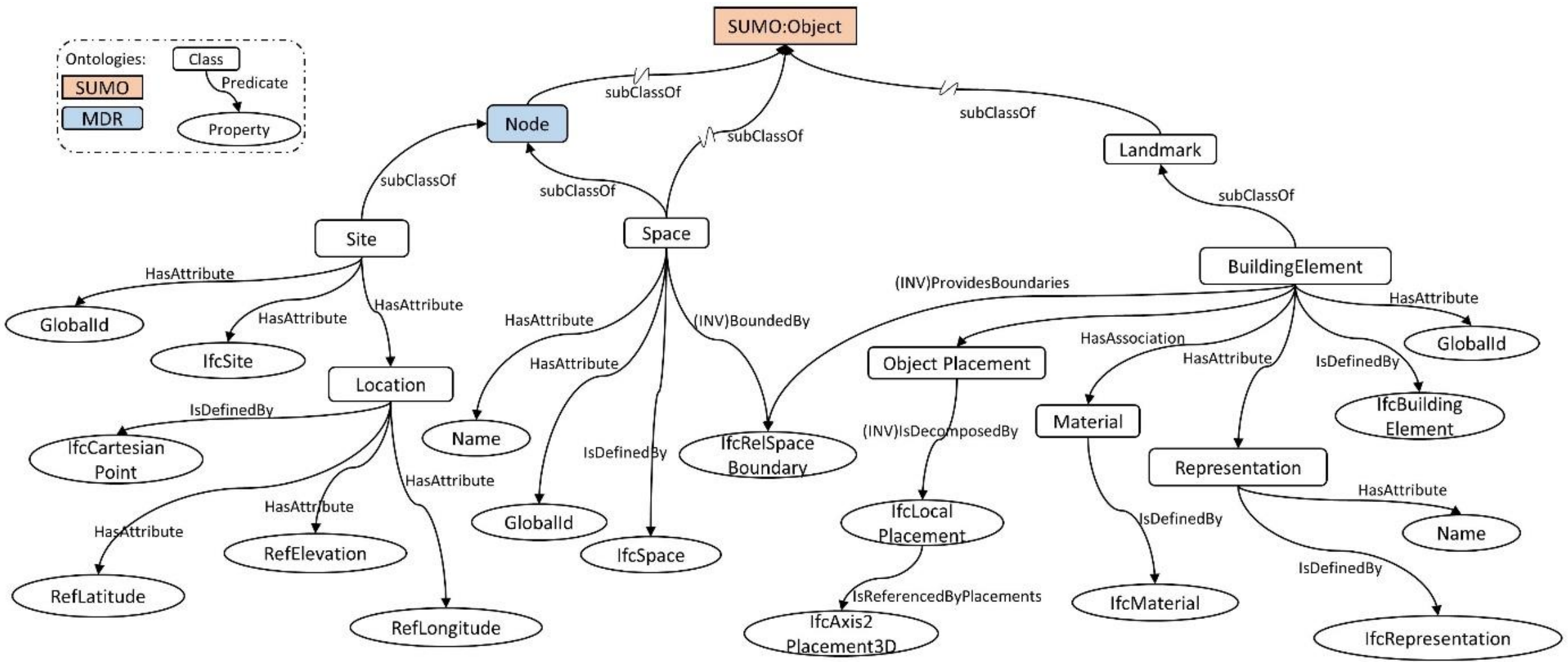

*FIG. 4. Partial indication of IFC classes relationships in BIRS ontology.*

From the extracted data, a topological map with each *space* considered as a node is expected as the output. FIG. 3 illustrates the relationships between *space*, topological map and *CORA:Region*. In this respect, a *space* is a node that is bounded using the *IfcRelSpaceBoundary* relationship which itself provides boundaries for building elements. As illustrated in FIG. 4, *IfcRelSpaceBoundary* provides relationships with *landmarks* and nodes used to generate the topological map containing the semantic information of the *spaces* and the *landmarks* for the purpose of robot navigation (implemented in Python). The use of IFC semantics for robot navigation is twofold. It contains the information of the *landmark* material in case the robot gets into a *space* for which the laser provides a point cloud unable to segment. In addition, the information of the nodes which are representing space names would enable semantic navigation. The application of the ontology for outdoor environments with knowledge of obstacles helps the robot to avoid collision.

For the robot to gather information about surrounding of the building, existing and natural artifacts are integrated, namely: existing buildings, water surfaces and vegetation are considered as obstacles. Figure 5 illustrates the relationships between CityGML and the BIRS ontology. The most important properties of outdoor artifacts are their location. The geo-location properties of the landmarks are translated to BIM. Then, the integrated data are translated to the robot local position system to support obstacle avoidance. The procedure on how the GIS information is transferred to BIM is beyond the scope of the current study. More information in this regard can be found in (Karan Ebrahim P. et al., 2014).

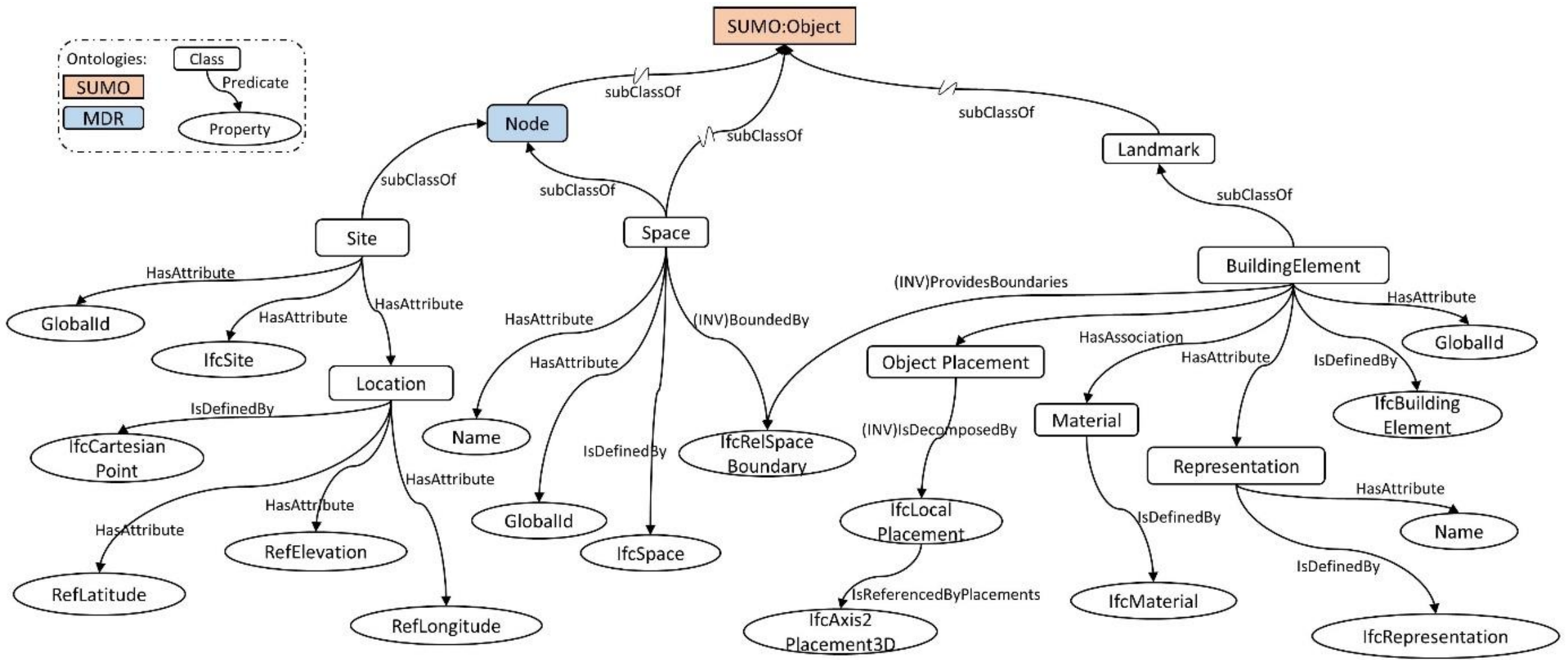

*FIG. 5. Partial indication of CityGML classes relationships in BIRS ontology.*

In the expected topological map, each pair of nodes are connected by an edge and since the edges are one-way relations, the resulting graph is directed. Extracted data from BIM, the edges are the transition between spaces (nodes), which is done through the doors (edges). *IfcDoor* is the *SubClassOf IfcBuildingElement* containing height and width which are defined through *IfcDoor* properties. Figure 6 illustrates the door relationships with regards to the landmark and to the edge. Furthermore, *IfcDoor* entity stores the information of door opening direction through y-axis of *ObjectPlacement*. BIRS uses *IfcDoor* information to create the edges. The center of the door in *IfcDoor* is defined by *IfcLocalPlacement* which inherits from its super-type *IfcProduct*. As it is illustrated in FIG. 6, a door is also a *landmark* with which the robot can localize itself. If the spaces are not connected through a wall opening, the coordinates of wall opening are extracted through *IfcLocalPlacement* of *IfcOpeningElement*.

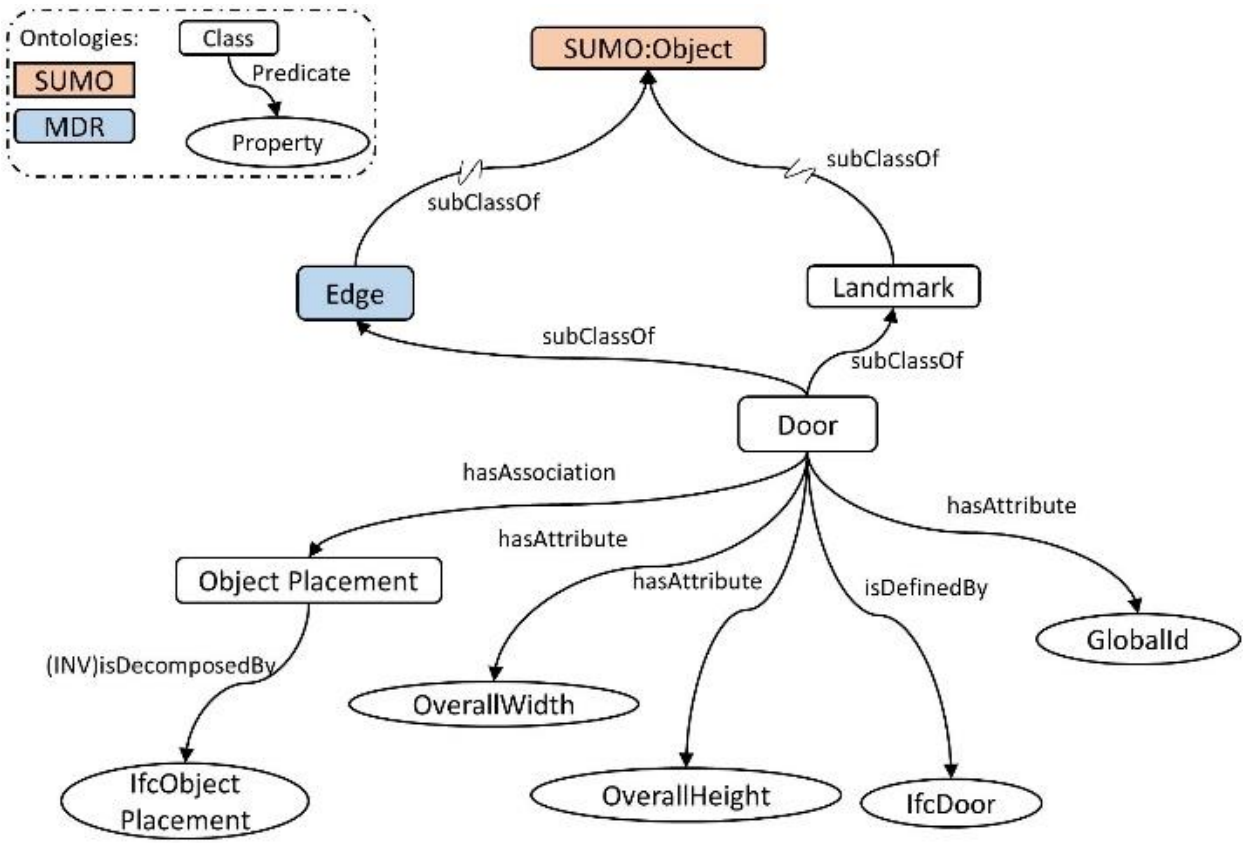

*FIG. 6. IfcDoor relationships in BIRS ontology.*

### 3.2. Building Information Robotic System (BIRS) Data Exchange

To achieve cross-domain data exchange using BIRS ontology, the IFC file is imported in Autodesk® Revit 2020 and processed with a Dynamo script (visual programming tool for Autodesk® Revit). The information for creating a semantic topological map is retrieved using the IFC parameters defined by the BIRS ontology. The correspondent information is then exported as XML database containing semantic information for creating the topological map. We then created the semantic topological map with the nodes and edges in a Python script. Since the data need to be accessible to ROS, this Python query script (ROS node) parses the BIRS information in the XML database. The semantic information from BIRS is not directly understood by the robot since the information is represented as strings with meaning in the ontology. Hence, the node includes dictionaries translating the complex information to machine-friendly scales. Apart from the rooms' names, all the information in the graph's nodes and edges follow the same translation process. Any other node in the ROS environment can then subscribe to retrieve semantic information. As an example, when the robot enters a room with a curtain wall hard to detect by its sensors, the BIRS node provides the robot with enough information about the room so it can rely on the BIRS layout (occupancy grid). The BIRS node can quickly find the room where the robot stands from the room boundaries exported automatically using the low-level information of the navigation system.

Due to the nature of the construction projects, there is a high number of activities and machinery which temporarily operate during the construction phase and are not fully represented in BIM. Hence, those entities that cannot be addressed with BIM are categorized under *Uncertainty* in the BIRS and are addressed with the low-level navigation stack and real-time sensors. The low-level navigation system leverages sensors on the UGV and the state-of-the-art algorithms to enable the robot autonomous navigation on construction sites. High-level information is provided to the robot from BIM-GIS data which gives the robot an *a priori* map (as discussed in section 3), as well as waypoints for navigation, localized building elements material and the information of nodes and edges in the topological map. All of the sensing modalities (e.g. laser scanner, odometry, depth camera and inertial measurement unit) used for navigation, path panning and obstacle avoidance are subject to various noise, such as process noise, model noise and electric noise generating *Uncertainty*. Detail on how the low-level navigation operates with the *Uncertainty* is beyond the scope of this paper.

# 4. CASE STUDY

Many use cases can potentially benefit from BIM-GIS integration into robot navigation stack. The selected use cases aim to exemplify the application of using BIM and GIS to help the robot understand the spatial structure and the topographical environment in an autonomous navigation stack on construction sites. The context of the case is pavilion D (2922.67 $m^2$) of ÉCOLE DE TECHNOLOGIE SUPÉRIEURE (ÉTS), which is covered with very light vegetation and surrounded by streets on two sides (Southwest and Southeast) and by buildings on the other sides. As illustrated in FIG. 7, a topological graph database is created for the second floor of the ÉTS, pavilion D with BIM/IFC information using the BIRS ontology. Each node includes a set of information identified and extracted using the BIRS ontological approach. We then take BIRS information in order to create a second map supported by ROS, the *a priori* metric map (occupancy grid). The following four use cases are practical deployments tested with a Clearpath Jackal mobile robot.

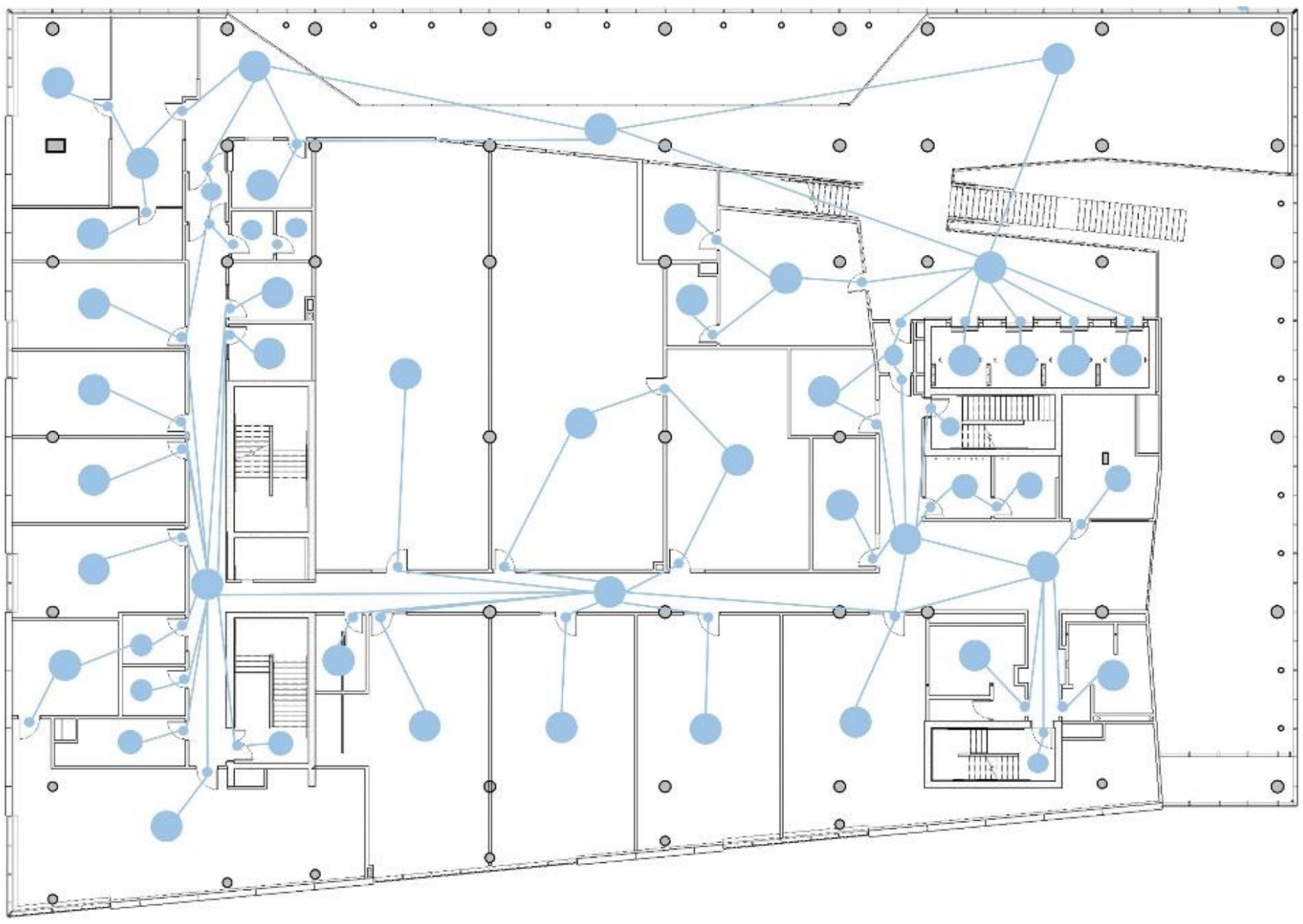

*FIG. 7. Topological Map representation of the ÉTS Pavilion-D.*

## 4.1. Semantic Indoor Navigation

In this use case, the robot starts navigation and data collection from "CORRIDOR OUEST 2019" (west hallway) and is instructed to reach "W.C. HOMMES 2002" (men bathroom) in the building. The destination room, on the Eastern section of the building, needs to be scanned while the starting point of the robot is on the Western section. The coordinates of the desired room (center of the room) are provided by the topological map to the robot in order to reach the destination. The query script provides the destination coordinates in the BIRS node and the low-level navigation system subscribes to the node and fetches the required information. As illustrated in FIG. 8, there is a long path to the destination with multiple rooms in between, namely, "VESTIBULE 2043", "HALL 2044", "VESTIBULE 2042", "CORRIDOR EST 2007", and "ESPACE CLLABORATIF 2004". On its way to the destination node, the robot enters "HALL 2044" which has a large and high wall, made of glass, invisible to the robot's sensor. Figure 8 illustrates the map created by the SLAM algorithm which used laser scanner for navigation.

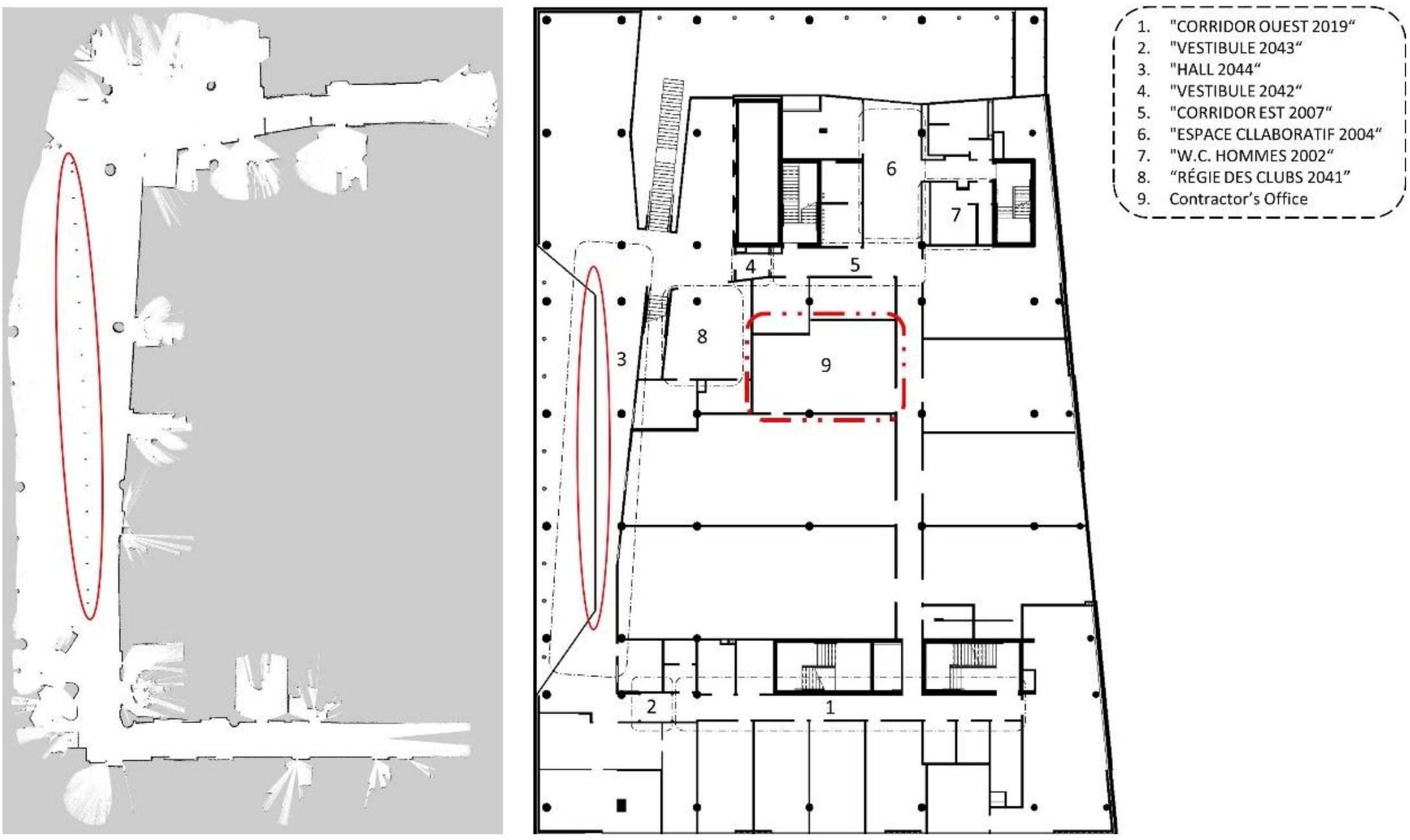

*FIG. 8. Map generated by SLAM algorithm (left) and map created from BIM (right).*

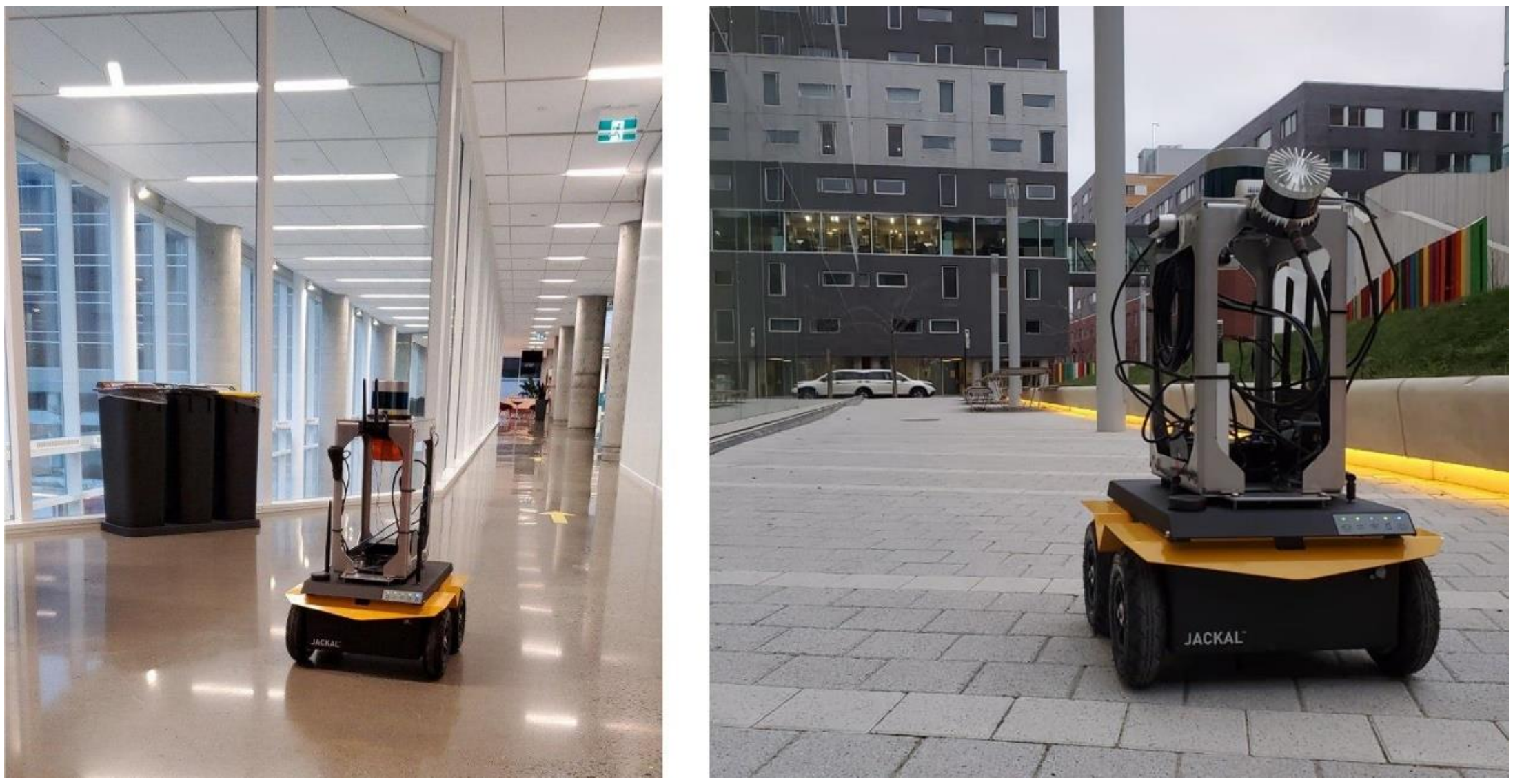

*FIG. 9. The curtain wall in "HALL 2044" (left) and the outdoor environment (right).*

It is shown that the curtain wall is not detected in the navigation system (see left part of FIG. 8 and FIG. 9). Before entering the room, the navigation stack fetches information to BIRS node about the next room. The material of that wall is listed as invisible to the sensor, so the navigation can rely on the BIM original layout instead of the robot’s sensor thereby, preventing the robot to collide with the wall. Then, for passing through "VESTIBULE 2042", the robot's sensors are able to detect the door entrance. However, the information of both doors for "VESTIBULE 2042" node such as door width, door height, and the central point of the door is also provided to complement the real-time sensors data. The robot's task is to get into the room "W.C. HOMMES 2002" to scan the room and monitor if the equipment is installed. In the destination node (room) the robot calls the database to query which equipment is to be installed so that the data collected can be compared with the as-planned model.

### 4.2. Semantic Outdoor Navigation

Another case study scenario relates to robot navigation in outdoor environments. In this use case, the robot navigates on the site around the building to collect data while there are several obstacles to be detected through BIM-GIS in order to contribute to obstacle avoidance. There is a landscape polygon covered with light vegetation, North of the building which is between the building of Pavilion-D and another building of the ÉTS facility. The polygon shape of the artifact contains 9 vertices. The coordinates of each vertex are retrieved from GIS and translated to the BIM environment. Similarly, the location of the vertices is provided by the BIRS node so that the polygon forms as an obstacle for the navigation system. Furthermore, outer boundary of the building is a wall, made of glass which the robot can hardly detect properly (see right part of FIG. 9). As shown in FIG. 10, all the information is encoded within an occupancy grid map compatible with ROS. The BIRS node provides the robot navigation system with all the required information; so it successfully complements the low-level information.

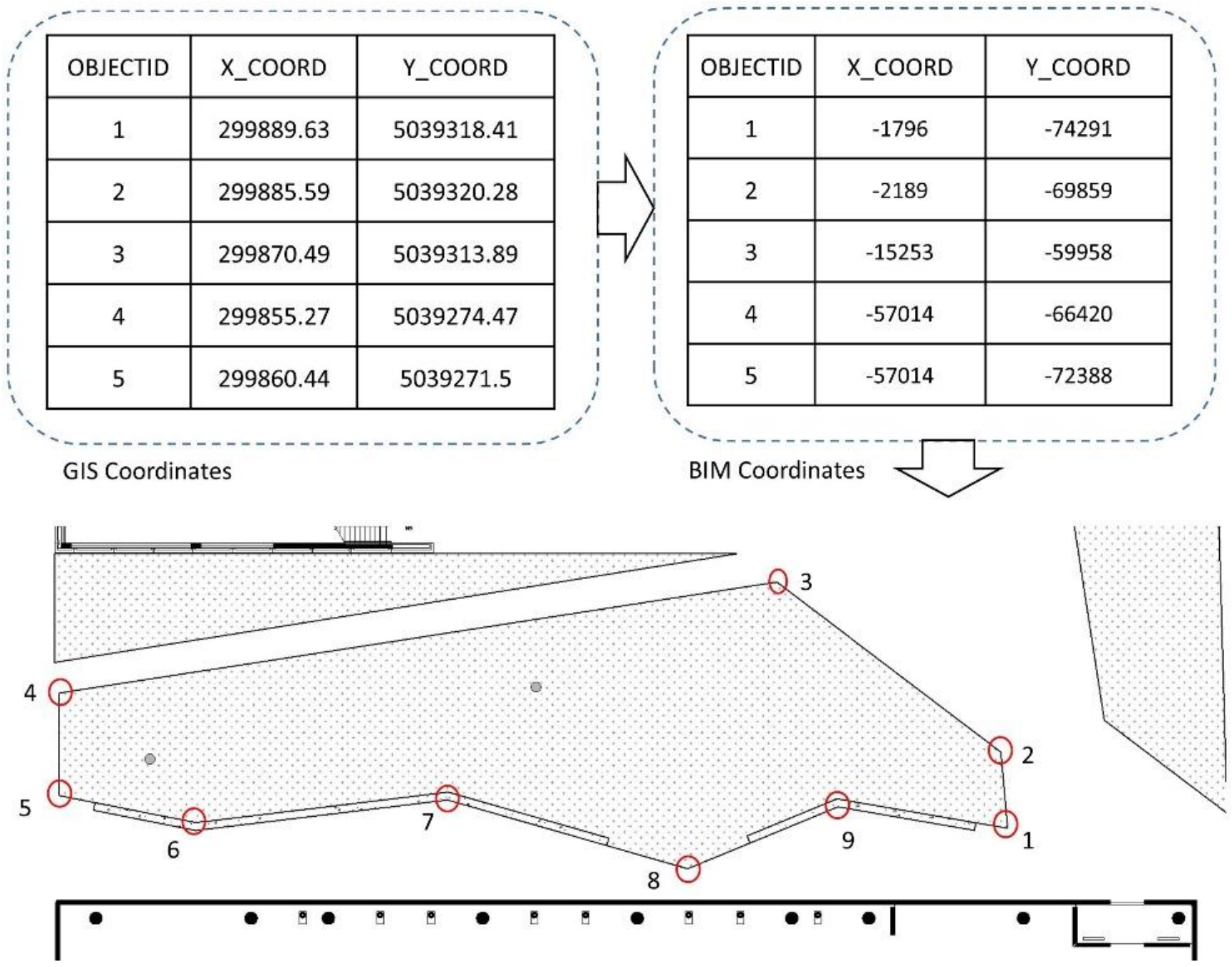

*FIG. 10. Partial indication of coordinate transfer from GIS to BIM which creates an occupancy grid map.*

## 4.3. Progress Monitoring

In the third scenario, while the robot is navigating the construction site, it enters “CORRIDOR SUD 2010” (South hallway). As illustrated in FIG. 11, the robot generates a map (see left part of FIG. 11) which is the as-built map (real state) of the section covered by the robot. In this case, the as-planned map (see right part of FIG. 11) shows no planned wall at the time of data collection. This difference is not a sensors default; it shows that the project is ahead of schedule and the two walls on each side of the hallway were already installed. With the query script that was developed in the BIRS ontology, the robot has access to the BIM data and it fetches the information in this regard. There are two uninstalled walls defined as *landmarks* as *SubClassOf CORA:Region* (*SubClassOf SpatialStructureElement*) in the BIRS ontology and their correspondent information derived from IFC. As elaborated in the BIRS ontology, the wall on each side of “CORRIDOR SUD 2010” is a *landmark*; a building element defined by IfcBuildingElement. The correspondent information such as the GlobalID, location, material and room boundary relationships defined in the BIRS ontology is available through the BIRS node.

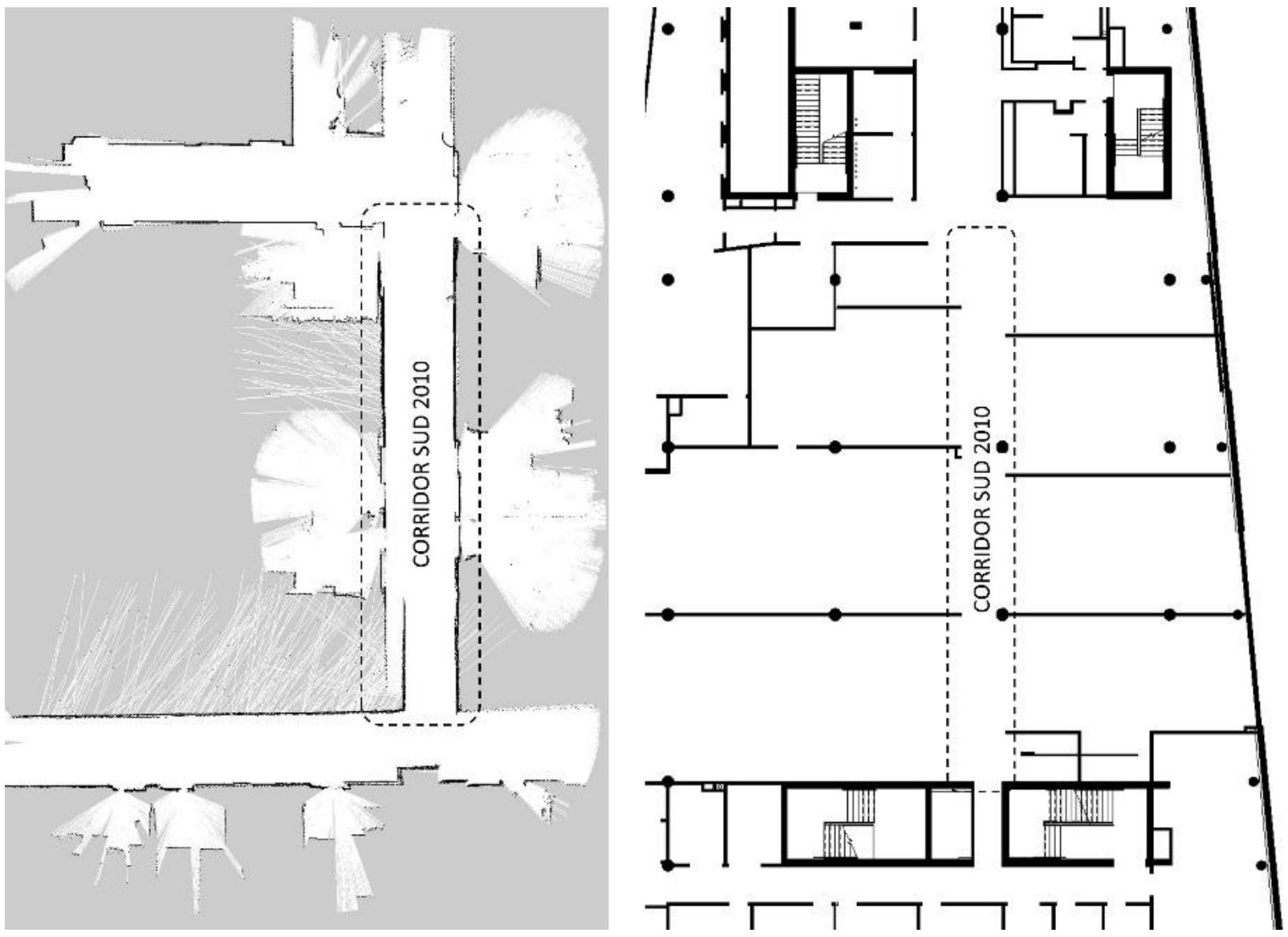

*FIG. 11. The As-built map created by the robot (left) and the As-planned map generated from BIM (right)*

Based on the information retrieved from BIRS, the robotic system identifies which walls are installed ahead of schedule by having all the relevant details provided by the BIRS ontology. Similarly, this process can be followed for all the building elements which ultimately enables detailed and accurate comparison between as-planned and as-built.

## 4.4. As-planned vs. As-built

The last use case scenario illustrates the comparison between as-planned and as-built on site. In this scenario, when the robot is collecting data, it detects anomaly which requires assistance. While the robot is on its way from "CORRIDOR OUEST 2019" to "W.C. HOMMES 2002" to collect data - right before entering "VESTIBULE 2042" - the robot detects a column as a *landmark* that is not planned in that location. Figure 12 illustrates the as-planned building layout (see right part of FIG. 12) and the detected location of a column in the as-built map generated by the robot (see left part of FIG. 12). The BIRS ontology provides the information to the robot that no column is planned in that location. In this circumstance, an anomaly is detected on "NIVEAU 2" and the robotic system finds the closest contractor's office to seek assistance. As illustrated in FIG. 12, the BIRS ontology provides the robot with the location of contractor's office. The room's location is queried from the BIRS, so the robot can navigate itself to the office.

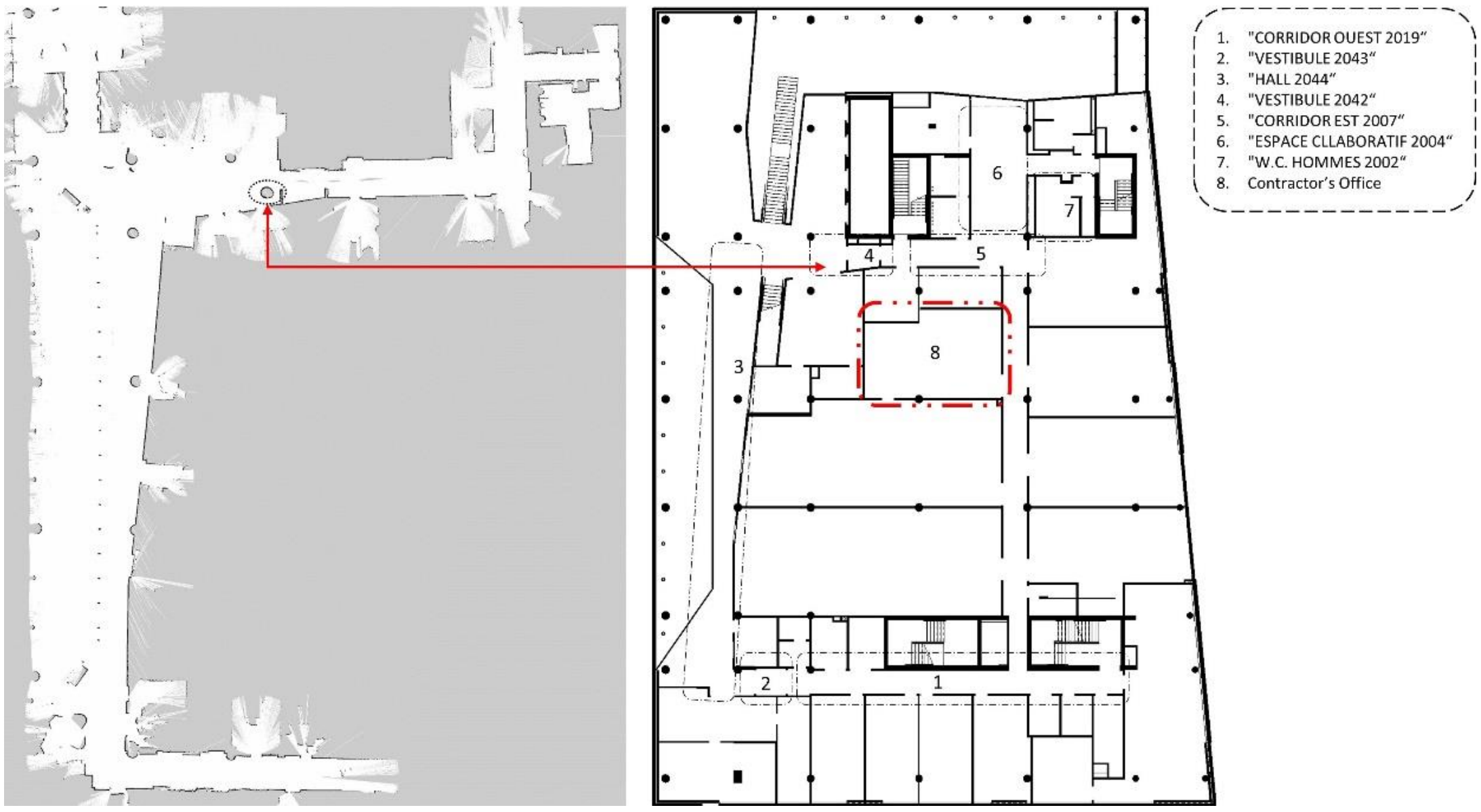

*FIG. 12. The detected anomaly on the map by the robot (left) and the as-planned map generated from BIM (right)*

The robot changes its path towards the contractor's office using a common warning signal to indicate that an anomaly was detected. This does not necessarily mean that the column is installed in an incorrect place and that is why the robot seeks assistance. This may happen when there is an obstacle in the occupancy grid map (generated by the robot) which is not in the map created from BIM. In this use case scenario, the detected anomaly is reported to the contractor for further analysis. A remote warning could also be sent if an operation room exists to supervise the robot's activities.

# 5. CONCLUSIONS

The current study is part of a larger research project aiming at the deployment of automated data collection and progress monitoring of construction projects. Even if several studies address robot navigation, the problem of semantic data translation from BIM-GIS to the robot navigation system was unsolved. Without a standard data format to translate data between BIM-GIS and ROS, the building semantics cannot be leveraged by any robot navigation system. Therefore, this study presented semantic data exchanges between construction and robotics. To achieve this goal, an ontology-based approach is adopted bridging the knowledge representation of building-related data to two well-established ontologies namely IEEE 1872-2015 and IEEE 1873-2015 standards; our first theoretical contribution. Then, the BIRS ontology enables cross-domain data exchanges between both domains and, ultimately, the BIRS is implemented in four use cases on a construction site deploying autonomous robot navigation. The process is done through defining an intermediate level in the BIRS to bridge the knowledge representation. IFC and CityGML semantics are first stored in topological database, parsed (through the script developed in this study), and translated to ROS for the robot navigation stack to be understood and be utilized. A second type of data format leveraged is the occupancy grids common to the robotic domain that is generated from the construction data and is used in the full navigation system. Four use cases were conducted to test the application of the proposed ontology. The current paper's hypothesis were: (1) autonomous data collection using mobile robot can help in automated progress monitoring and (2) autonomous robot navigation can be improved by leveraging building-related data. The results show that the ontology provides semantic data that helps the robot to navigate and to understand what information is expected from its sensors which helps acquiring accurate and more reliable data.

The BIRS applications can be extended to other construction areas such as calculating the productivity of the trades on site, creating updated circulation map for mobile equipment, creating evacuation plans for safety analysis. For example, the comparison between as-planned and as-built of a construction project can be made through comparing the map derived from BIM and the map generated by SLAM. Also, the information of Mechanical, Electrical, and Plumbing (MEP) equipment can be added to the ontology so that the data collected from the construction scene can be better compared to the as-built plan. In this direction, the robot can navigate the jobsite and collect data on a regular basis, thus providing information to the construction stakeholders on the productivity on the construction project. While the robot is navigation the construction site, it collects data through its sensors and generates a map of the environment through SLAM. The SLAM-generated and the BIM-driven maps complement each other and create an accurate map of the environment providing the location of the building elements and the temporary equipment and machinery on the site. The combined map can then be used for generating circulation map for the construction site. In addition, this map can also be leveraged for creating evacuation plans during the construction phase based on the existing conditions.

The current research takes *IfcBuildingElement* entity and its sub-classes into consideration to provide the robot with landmarks for navigation. However, the MEP equipment information is not included in the current study, i.e. if there is an equipment (obstacle for the robot), it would not appear on the map generated from BIM. This limitation can be addressed through extending the current ontology in order to contain MEP information in the BIRS ontology following the same methodology proposed in this paper. As it is shown in section 4.3, one of the use cases of the BIRS is progress monitoring. Including the MEP information of the building would provide more accurate and thorough progress monitoring of the construction project. The other limitation of the study is that the ontology aims at incorporating high-level semantic information of the building, while the practical robot navigation requires low-level information provided by the sensors and the strategies to implement the low-level navigation. The topological map developed in this paper is a rich database of building elements semantics that can be used for semantic path planning for the robot navigation. As the robot navigation is highly dependent on low-level sensor-based navigation, the authors aim to develop a strategy which uses the topological map and the sensors information to implement a full stack semantic navigation of the mobile robots on construction sites.

## 6. ACKNOWLEDGEMENT

The authors are grateful to Natural Sciences and Engineering Research Council of Canada for the financial support through its CRD program 543867-2019 as well as to Pomerleau - the industrial partner of the ETS Industrial Chair on the Integration of Digital Technology in Construction.